**Title:**

# Inferring continuous cognitive traits from limited neuroimaging data using pairwise convolutional neural networks

**Authors and Affiliations:**

**Marjan Rashidi[1], Robert Bain[2], Craig Stark[2], Jeffrey Krichmar[1] , Mary Hegarty[3], Elizabeth R. Chrastil[1,2,*]**

**1. Department of Cognitive Sciences, University of California, Irvine**
**2. Department of Neurobiology & Behavior, University of California, Irvine**
**3. Department of Psychological & Brain Sciences, University of California, Santa Barbara**

*** To whom correspondence should be addressed**

**Author Note**

**Elizabeth R. Chrastil https://orcid.org/0000-0003-2544-0152**

**Abstract**

Individual differences in cognitive abilities vary widely, yet the structural brain features supporting these differences remain poorly understood. Traditional brain-behavior analyses rely on predefined metrics such as volume or cortical thickness and typically assume linear relationships with behavior, limiting sensitivity in small datasets. Here, we introduce a pairwise convolutional ranking framework that learns directly from whole-brain T1-weighted MRI by making comparative rather than absolute predictions, which performs well even with small samples (N=108). We tested this method with four small-scale spatial abilities tasks. Regression analysis predicted performance with high accuracy across all tasks, achieving greater predictive power over standard regression and pointwise deep learning baselines. Iterative cortical-region removal (backward elimination) revealed brain regions supporting individual performance common to all four tasks, as well as task-specific regions. These results demonstrate that pairwise learning extracts meaningful behavioral information from structural MRI, providing prediction and interpretable insight into the neuroanatomy of individual differences.

## Introduction

The relationship between individual structural variation in the human brain and behavioral performance differences remains a central challenge in cognitive neuroscience. Despite decades of work relating brain structural features to behavior, brain-behavior associations are often modest, difficult to replicate, and sensitive to modeling assumptions, particularly in small samples [1]. Conventional structural MRI analyses typically relate cortical thickness, surface area, or regional volume to behavioral scores using linear models or region-wise correlations [2], [3], [4], [5], [6], [7], [8], [9], [10], [11]. While informative, these approaches generally explain modest variance [1], are sensitive to interregional collinearity, and assume that structure–behavior relationships are globally linear. Such constraints may obscure the distributed, network-level substrates supporting individual differences in cognition.

Deep learning models have improved behavioral prediction in large MRI datasets [12], [13], [14], yet these approaches often require thousands of subjects, treat the brain as a single 3D voxel tensor, and can be difficult to interpret neuroanatomically. There remains a need for predictive frameworks that are more data-efficient and capable of identifying anatomically meaningful contributors to behavior. We address this need using a ranking-based convolutional neural network (CNN) framework that reframes behavioral prediction as a pairwise comparison problem. Traditional models regress absolute scores, for example, a person got 72% of trials correct. In contrast to that approach, here the model learns to rank individuals, for example whether person 2 scored higher than person 1, thus increasing the number of effective training constraints and improving robustness in small datasets.

Here, we apply this pairwise CNN framework to predict visuospatial abilities from T1-weighted MRI in a sample of N=108 individuals. Instead of regressing behavioral scores directly from absolute performance, the model learns the relative rankings of participant performance [15]. We can then derive the predicted absolute score on a continuous scale using maximum entropy methods. This formulation increases the number of usable training signals, reduces sensitivity to score noise, and supports robust prediction in small datasets. To localize anatomical drivers of performance, we pair this approach with an iterative backward-elimination procedure that quantifies the impact of removing each cortical ROI while keeping feature representations fixed.

We used this approach to test four small-scale spatial abilities tasks— including object manipulation tasks (mental rotation, paper folding), and perspective taking tasks (viewpoint transformation, road map)— which are central to navigation, STEM achievement, and everyday functioning [16], [17]. Spatial abilities vary widely across individuals and depend on distributed neural systems spanning parietal, occipital, and frontal cortices [18], [19], [20]. A longstanding challenge in cognitive neuroscience is determining whether and how structural differences across the cortex predict meaningful variation in these visuospatial skills. Critically, prior neuroimaging work on these tasks has been dominated by functional imaging studies of regional activation, while structural brain correlates of individual differences in performance remain comparatively understudied.

We chose these four tasks expecting that some overlapping brain regions would emerge because of the common core of spatial abilities, while task-specific regions could pull apart object manipulation from perspective taking. Mental rotation (MRT) and paper folding (PFT) share highly similar object-based transformation demands and are typically strongly correlated in behavioral studies [21]. Viewpoint transformation (iVTT) is more distinct because it is implemented in an immersive VR environment that allows participants to use vestibular and proprioceptive cues in addition to visual imagery. The road map task (RMT) falls between these extremes; it primarily involves perspective-taking processes but also shares some overlap with object-based transformation mechanisms. Together, this range of tasks spans closely related and more differentiated spatial processes, providing a useful test case for distributed brain–behavior mapping.

Using multiple validated spatial measures, we show that this framework achieves substantially higher discriminability than classical regression and standard deep learning baselines. ROI-level backward elimination reveals a distributed set of predictive regions — including superior parietal lobule, precuneus, and precentral — consistent with established models of mental transformation, visuospatial attention, and object manipulation. Together, these findings provide a data-driven account of the structural substrates supporting individual differences in spatial ability and demonstrate how ranking-based predictive modeling can yield stable, anatomically interpretable insights in cognitive neuroscience.

## Results

### Experimental Tasks

We assessed multiple forms of spatial cognition—including mental rotation, paper folding, viewpoint transformation, and road map —to capture both object-based and perspective-based transformation processes (Figure 1). Mental rotation, paper folding, and road map were administered as desktop tasks, whereas viewpoint transformation was implemented in a fully immersive virtual reality (VR) environment. Among these tasks, mental rotation (MRT) and paper folding (PFT) primarily index object-based manipulation, requiring participants to mentally rotate three-dimensional objects. In contrast, viewpoint transformation (iVTT) and road map (RMT) emphasize perspective taking, involving imagined rotations of the self within space rather than rotation of external objects. Task details are provided in the Methods.

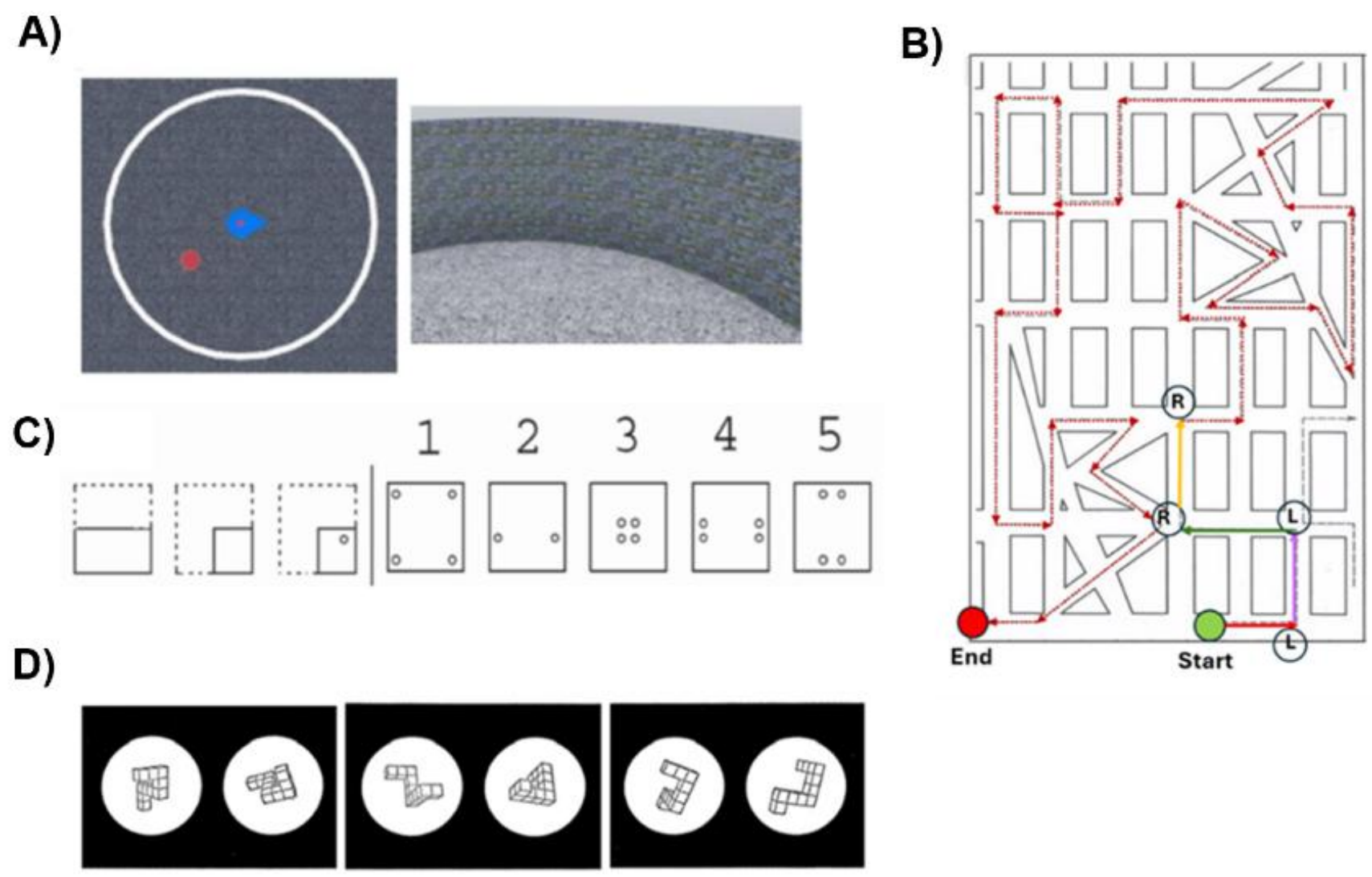

**Figure 1. Experimental Tasks**. **A)** Immersive Viewpoint Transformation Task (iVTT; perspective taking). Overhead map view shows the position and orientation of the participant (blue) and the location of the goal (red) in the fully immersive and ambulatory virtual environment. The location of the circular arena is shown in white. The participant then walks in a virtual desert with a walled arena to where they think the goal was located. **B)** Road Map Task (RMT; perspective taking). Red dotted arrows indicate the routes to be taken by participants. The correct answers for left and right turns are marked with "L" and "R" for the first four turns, only for illustrative purposes. For each corner, participants must decide whether each turn is a left or right turn without rotating the map. **C)** Paper Folding Task (PFT; object-based). The left figure shows a piece of paper folded multiple times, with holes punched through all the folds. Participants must then choose from five options on the right side to identify which unfolded paper matches the result of the hole punch. **D)** Mental Rotation Task (MRT; object-based). For each pair of figures, participants must decide whether the figure on the right is a rotated version or a mirror-image version of the figure on the left.

### Pairwise Supervision Expands the Effective Dataset Size Quadratically

A central challenge in structural MRI prediction is the extremely limited number of subjects available for supervised learning. To address this problem, we reformulated behavioral prediction as a pairwise ranking problem, where each training example consists of two brain scans and a binary label indicating which subject performed better. Rather than training on the absolute performance score (e.g., 75%, 82% correct), the training instead relies on relative comparisons (e.g., person 1 performed better than person 2). This transformation converts a dataset of $N$ subjects into $N(N-1)/2$ unique training comparisons, yielding a quadratic expansion in available data relative to classical single-subject regression. For our sample of 108 participants, this method produces 4,950 unique ordered pairs before augmentation (e.g., left–right flips across the sagittal plane as the primary augmentation; see Methods) effectively transforming a small-sample regime into one with thousands of balanced training examples. Figure 2 illustrates how the number of available comparisons grows quadratically with sample size, allowing even modest datasets to generate a rich and diverse training signal. This combinatorial expansion forms the statistical foundation that enables our model to generalize despite the limited number of individual participants.

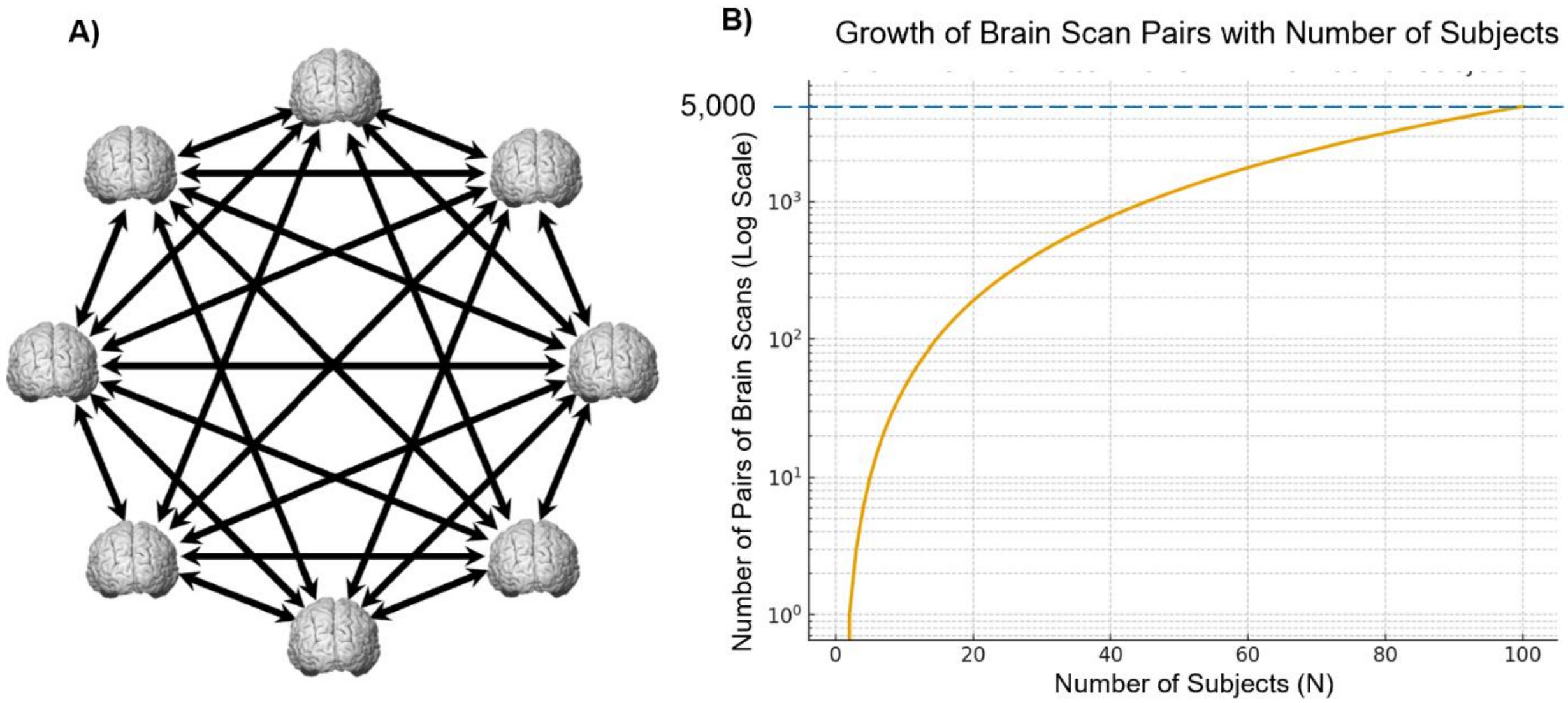

**Figure 2 | Pairwise supervision transforms small neuroimaging datasets into large relational training sets.**
**A)** Fully connected pairwise comparison graph illustrating the relational structure among subjects. Each brain represents one participant, and each directed arrow denotes a binary decision ("Which participant performs better?"). This forms a complete directed graph over the cohort, generating a training signal from every ordered participant pair. **B)** Quadratic growth of unique participant pairs as a function of sample size. A dataset with $N$ subjects yields $N(N-1)/2$ distinct unordered pairs (plotted on a log scale), enabling thousands of training examples even for small sample sizes. For example, 108 subjects produce 4,950 unique pairs before augmentation. Together, panels (A) and (B) illustrate how pairwise ranking converts limited subject-level data into a rich, high-volume supervisory signal that supports stable and data-efficient deep learning in small-sample neuroimaging datasets.

We used a pairwise convolutional ranking model that learns to compare two structural MRI scans and decide which individual exhibits higher spatial ability, using standard cross-entropy loss. The architecture consists of a shared slicing network that extracts subject-level features from whole-brain T1 volumes and a ranking network that produces the pairwise decision (Figure 3). This framework allows us to derive robust subject rankings and perform backward regional-elimination analyses to identify cortical contributions to spatial ability.

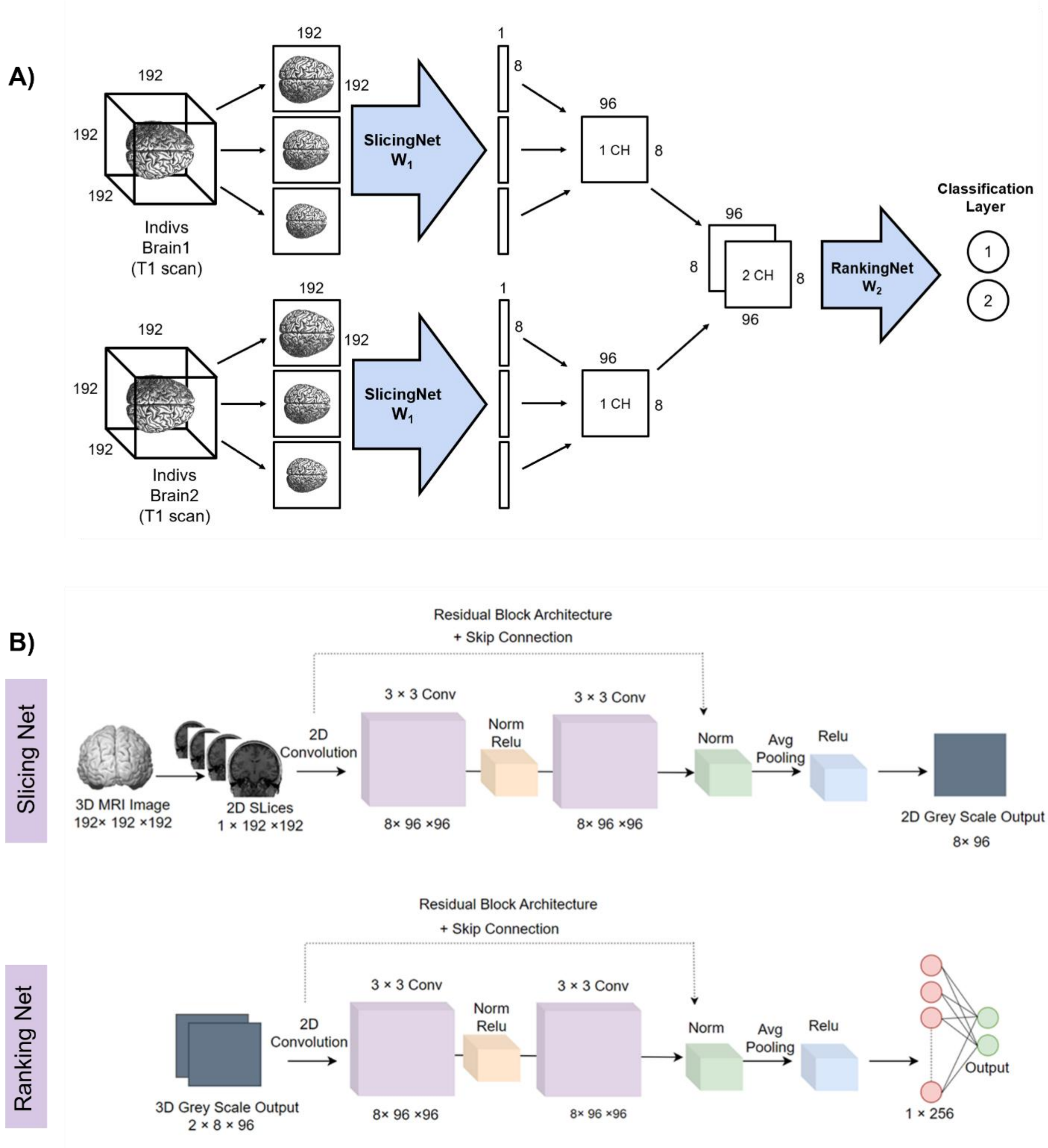

**Figure 3 | Architecture of the backward-elimination pairwise (bep) model.**
**A)** High-level overview of the pairwise convolutional neural network (CNN) framework. A 3D T1-weighted MRI volume is decomposed into 2D slices and passed through a slicing network. Two participants are processed independently using shared weights ($W_1$). Their resulting feature representations are concatenated along the channel dimension and passed to a ranking network ($W_2$), which outputs a binary decision indicating which subject shows higher behavioral performance. **B)** Detailed view of the same pipeline. The top row illustrates the slicing network applied to a single subject: 2D slices extracted from the 3D input are processed through residual blocks (one block shown for simplicity), producing a subject-level 2D feature map (1 channel). The bottom row shows the pairwise stage: 2D feature maps from two subjects are stacked along the channel dimension (forming a 3D 2-channel input) and passed through the ranking network, which also consists of residual blocks (one block shown). The final fully-connected head produces the binary prediction. For clarity, only a single residual block is depicted in each network; in practice, the slicing network contains multiple blocks (i.e., 5), and the ranking network contains fewer (i.e., 2).

**Deriving Continuous Spatial Ability Scores from Pairwise Comparisons**

To estimate continuous behavioral scores at test time from a model trained on pairwise comparisons, we used a reference-based inference procedure. The reference set comprised all labeled participants in the dataset, including participants assigned to the held-out test subset. In this approach, an unlabeled sample was compared against multiple labeled references, and its value was inferred by aggregating those pairwise outcomes (See Supplementary materials Figure 1). For reference scores represented by multiple participants, the predicted probabilities were averaged to obtain one probability estimate at each observed score. This formulation provided each example with richer supervision than conventional pointwise regression and helped avoid the low-variance collapse often seen in small-sample settings (e.g., Figure 4A). Finally, to obtain individualized estimates of continuous spatial performance from the pairwise model, we compared each test participant against all individuals in the reference set and recorded the network's predicted probability that the test participant would outperform each reference subject.
This procedure produces two complementary signals:

1. An inflection in the probability curve, where the model switches its relative judgment from better performance to worse performance (or vice versa) than the reference subject, and
2. A peak in decision entropy, reflecting maximal uncertainty when the test and reference subjects are most similar in underlying score.

The score at which entropy is maximized provides a stable and data-driven estimate of the participant's continuous spatial ability. Figure 4B illustrates this inference process for MRT for one participant, demonstrating how the pairwise CNN yields continuous regression-like outputs without requiring an explicit regression head.

***Population-Level Regression Using Pairwise Comparisons Outperforms Standard Models***

To test the pairwise model's prediction performance, we tested it against several current state-of-the-art models. We first tested a direct pointwise CNN regression as a baseline for comparison. The baseline was matched closely to the pairwise architecture: it used the same slice encoder, the same $192^3$ MRI input, and the same 2D slice-based processing pipeline with augmentation, but operated on a single scan rather than a pair of scans and predicted the target score directly through a linear output layer optimized with mean squared error. Figure 4A presents true versus predicted MRT scores for the this baseline. Across subjects, predictions concentrate within a narrow band near the training-set mean, showing limited dynamic range and failing to reflect subject-specific variation, with a training mean absolute error (MAE) of 0.848 and test MAE of 0.924. Similar behavior was observed for feature-based baselines (see Supplementary Materials Figure 2). Ridge regression and Elastic Net, applied to cortical thickness features, also failed to recover subject-level variability, producing compressed predictions with low dynamic range and substantial error (Ridge: test MAE = 0.874;  Elastic Net: test MAE = 0.924). Although these models used standard region-level summaries, their predictions remained biased toward the mean and did not capture the full range of individual differences in MRT performance.

To evaluate the accuracy of these continuous predictions using pairwise framework, we compared inferred behavioral scores against ground-truth standardized MRT performance across all participants (Figure 4C). Predicted scores closely followed the identity line, with a Train MAE of 0.348 and a Test MAE of 0.647. This result demonstrates that relative-comparison supervision can be reliably transformed into meaningful continuous behavioral estimates, despite the absence of any explicit regression loss during training.

Behavioral score predictions for the other tasks are shown in Figure 4D–F. Across the other three tasks, training and test produced accurate predictions (PFT train MAE = 0.245, test MAE = 2.625; iVTT train MAE = 0.069, test MAE = 0.065; RMT train MAE = 1.071, test MAE = 3.125). In contrast, pointwise CNN regression baselines produced poor fits (PFT train MAE = 2.776, test MAE = 2.893; iVTT train MAE = 0.587, test MAE = 0.500; RMT train MAE = 4.487, test MAE = 4.212; see Supplementary Materials Figure 2). These results confirm that the pairwise ranking framework captures stable, subject-level variation in spatial ability across multiple tasks.

We also applied the same regression approach to the RMT task in an independent dataset, allowing cross-study comparison of the regions identified as predictive of performance (see Supplementary Materials Figure 4). This replication achieving a training MAE of 1.862 and a test MAE of 4.625. These MAEs are a bit larger than in the original dataset, but still substantially smaller than the baseline.

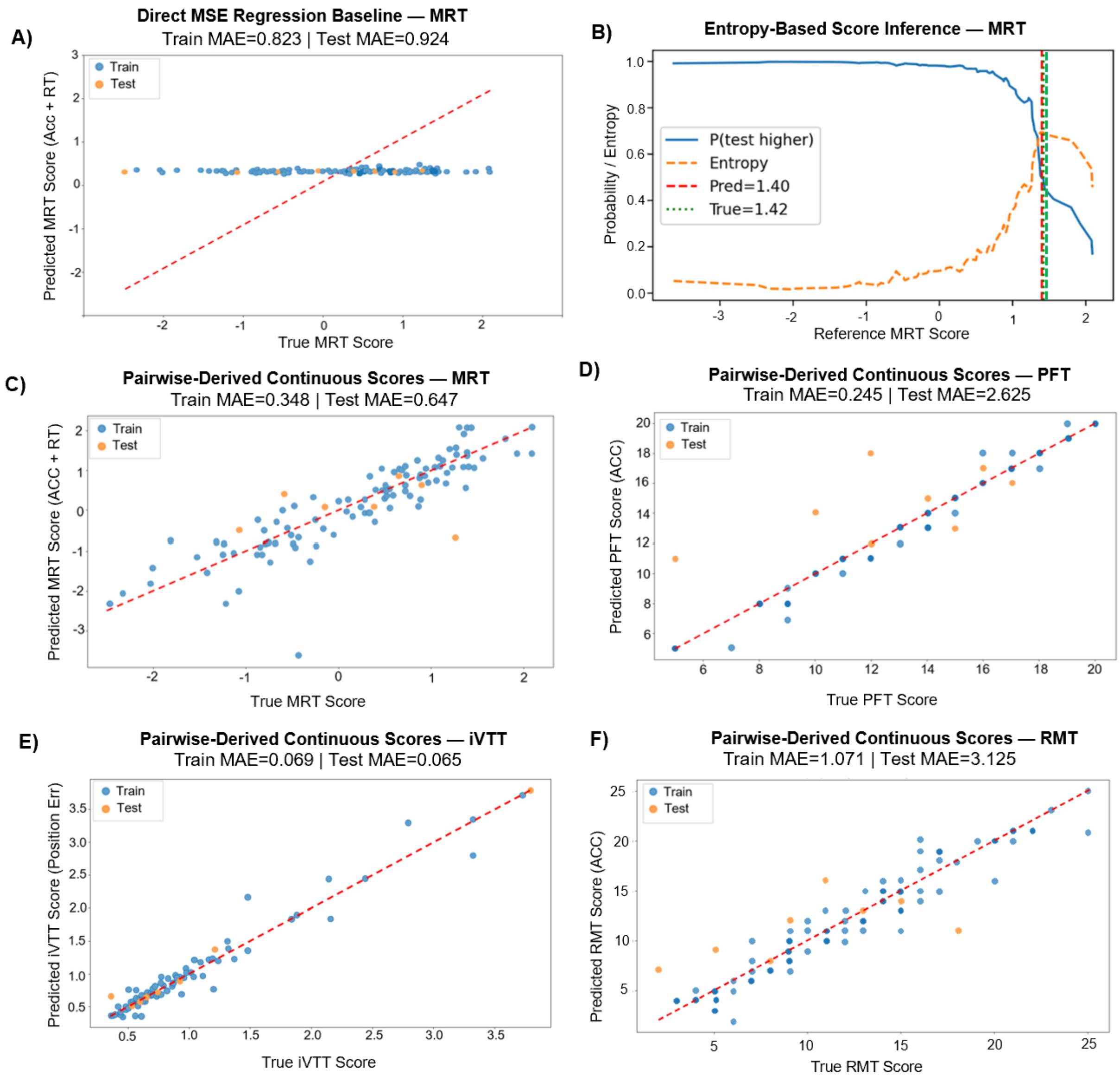

**Figure 4. Converting pairwise rankings into continuous MRT score predictions using an entropy-based regression procedure.**

**A)** True versus predicted MRT scores (z-score of accuracy and response time) for the direct mean squared error (MSE) regression baseline. Predictions are compressed within a narrow range and show limited alignment with the identity line (dashed red), indicating poor recovery of the full range of MRT scores (MAE = 0.924). **B)** Entropy-based regression procedure for the Mental Rotation Task (MRT). For each test participant, the pairwise model is compared against all reference participants across the MRT score range. The model outputs the probability that the test participant scores higher than the reference participant (blue curve) and the associated decision uncertainty (entropy; orange dashed curve). The predicted score is taken at the point of maximum uncertainty, where this probability is approximately 0.5, indicating that the model is least certain which participant has the higher score and that the two scores are most similar. The predicted score (red dashed line) is shown alongside the true MRT score (green dotted line). **C–F)** Continuous regression estimates derived from the pairwise ranking model for **C)** MRT (Train MAE = 0.348, Test MAE = 0.647), **D)** Paper Folding (PFT) (Train MAE = 0.245, Test MAE = 2.625), **E)** Viewpoint Transformation (iVTT) (Train MAE = 0.069, Test MAE = 0.065), and **F)** Road Map (RMT) (Train MAE = 1.071, Test MAE = 3.125). Predicted scores (y-axis) closely track true behavioral performance (x-axis) for both training (blue) and held-out test subjects (orange). The dashed red line indicates the identity line. Despite training only on pairwise comparisons—without any direct regression objective—the model generates accurate continuous predictions, demonstrating that pairwise supervision can be reliably transformed into individual-level behavioral estimates across diverse spatial tasks.

***Label-Shuffling Control Analysis***

To verify that the model captured meaningful brain–behavior structure rather than overfitting to noise, we performed a label-shuffling control. When trained with the true MRT scores, the network showed steadily decreasing loss on both the training and test sets, demonstrating successful generalization to unseen subjects (Figure 5A).

In contrast, when behavioral labels were randomly permuted, the model rapidly overfit: training loss continued to decrease, but test loss diverged toward chance performance (Figure 5B). This dissociation confirms that the strong predictive performance observed with the true labels reflects genuine neuroanatomical signal, rather than artifact, memorization, or spurious correlations.

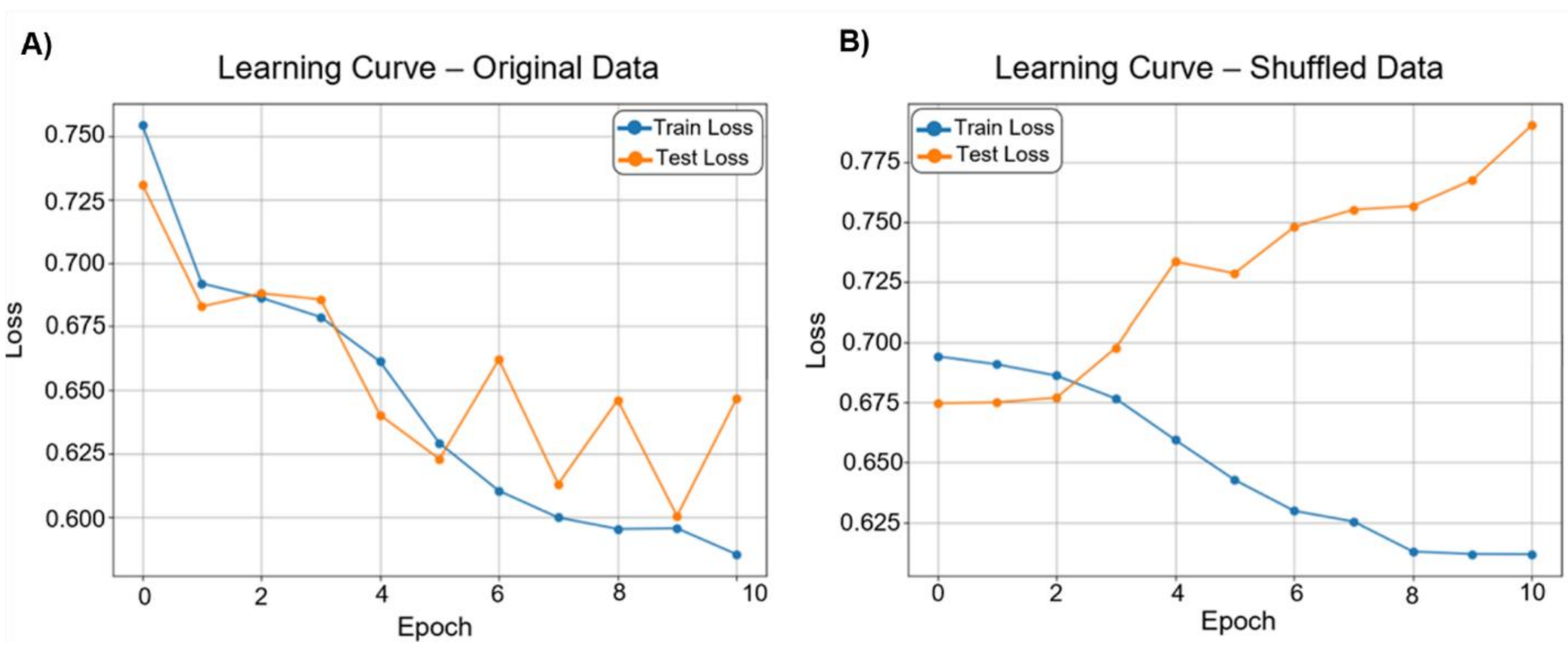

**Figure 5. Control analysis using shuffled behavioral labels.**
**A)** Training with correct mental rotation scores produces decreasing loss on both training and test sets, indicating successful generalization to unseen subjects. **B)** When behavioral labels are permuted, the model cannot learn meaningful structure in the MRI features. Training loss continues to fall while test loss diverges, demonstrating severe overfitting. This check confirms that the model's predictive performance in the real-label condition is not driven by subject identity or random noise.

**Backward ROI Elimination Reveals a Distributed Structural Basis for Spatial Ability**

To obtain anatomically-grounded interpretability, we applied systematic backward ROI elimination, which assesses how the removal of each cortical parcel affected pairwise prediction accuracy. Unlike voxel-level saliency maps which lack a well-defined biological ground truth in structural MRI and can be sensitive to model initialization [22], backward elimination provides a performance-based estimate of regional contribution by explicitly evaluating how predictive accuracy changes when each region is removed. The pairwise convolutional neural network (CNN) reliably learned to rank individuals by mental rotation performance using only their T1-weighted structural MRI scans. In the initial round of backward elimination, all 60 ROI-held-out sub-models—each trained while withholding a different cortical or subcortical ROI—achieved high test accuracy (around 90%) indicating strong generalization to previously unseen subjects (see Supplementary Materials Figure 5).

In each backward-elimination round, a full pairwise model was first trained on the current ROI set, after which the slicing network was fixed and cached features were used to train ROI-held-out ranking sub-models, one per remaining ROI. Eliminating individual ROIs produced reliable and interpretable accuracy drops, revealing a coherent spatial pattern across folds of the regions that make significant contributions to the model predictions. Figure 6 illustrates these regional contribution profiles, demonstrating that visuospatial ability depends on a distributed cortical network rather than isolated loci.

Across tasks, high prediction accuracy was maintained despite extensive regional masking, indicating that spatial abilities are supported by a distributed yet selective set of informative cortical regions rather than isolated loci. In contrast, training the network on randomly selected ROIs did not reproduce the elevated accuracy observed when the model was trained on the informative ROIs identified by backward elimination, supporting the validity of this approach (see Supplementary Materials Figure 6). Unlike conventional structural analyses that assess associations between cortical thickness or grey matter volume and behavioral measures one ROI at a time, backward elimination provides a multivariate estimate of regional contribution by measuring how prediction accuracy changes when each ROI is removed from the model.

***Common and Specific Contributions of Brain Regions across Tasks***

To assess whether different spatial abilities rely on shared structural features, we quantified the overlap between the top ROIs identified for each task (Figure 6A and 6B). Across all four tasks, five brain regions emerged as being highly influential for model prediction: superior frontal, rostral middle frontal, superior parietal, lateral occipital, and precentral cortices. However, the relative importance of these regions varied by task, as did the magnitude of the performance drop when they were removed. MRT and PFT also shared precuneus and fusiform, for a total of 70% overlapping regions. PFT and RMT shared inferior parietal, with 54% total overlap. Outside of the five core regions, iVTT did not share any overlapping regions, with 45.5% total overlap for all the other tasks. Notably, iVTT is the only task that involves physical movement and is the only task that showed postcentral (sensory cortex) involvement. In total, seven regions were unique to a single task (Figure 7A and 7B). It is also important to note that for some tasks such as mental rotation, backward elimination reduced model accuracy much more strongly (i.e., model accuracy down to 85.1% for the top ROI) than for other tasks (e.g., paper folding still had 92.6% accuracy after removing top ROIs).

Together, these results suggest that while all tasks draw on a distributed visuospatial network, object-based transformations (MRT and PFT) exhibit a more tightly shared cortical substrate, whereas viewpoint updating (iVTT) engages partially distinct structural systems supporting perspective taking. RMT was somewhat in the middle, suggesting that it shares elements of object-based spatial processing while also engaging distinct mechanisms. Although the task requires perspective taking, it is performed from a fixed, map-based viewpoint rather than an immersive VR setting, which may explain its intermediate similarity to both object-based and viewpoint-dependent tasks.

These tasks have a considerable amount of overlap in the brain regions supporting them, but they also share substantial cognitive features: they are all small-scale spatial abilities. To test whether our findings are an artifact of the method or whether there is true overlap in these regions, we conducted an analysis on a very different task – a self-reported scale of chronic stress – using the same method (Figure 7C and 7D; Supplementary Materials Figure 7). Although several cortical regions overlapped with the spatial tasks (e.g., superior parietal and lateral occipital), the emergence of the hippocampus as an influential region differentiated this analysis from the visuospatial tasks. This finding is notable given the well-established role of the hippocampus in stress regulation and its high density of glucocorticoid receptors [23]. The selective involvement of hippocampus in the stress analysis supports the specificity of our approach, demonstrating that the framework does not simply recover a fixed set of regions but instead identifies task-relevant structural substrates.

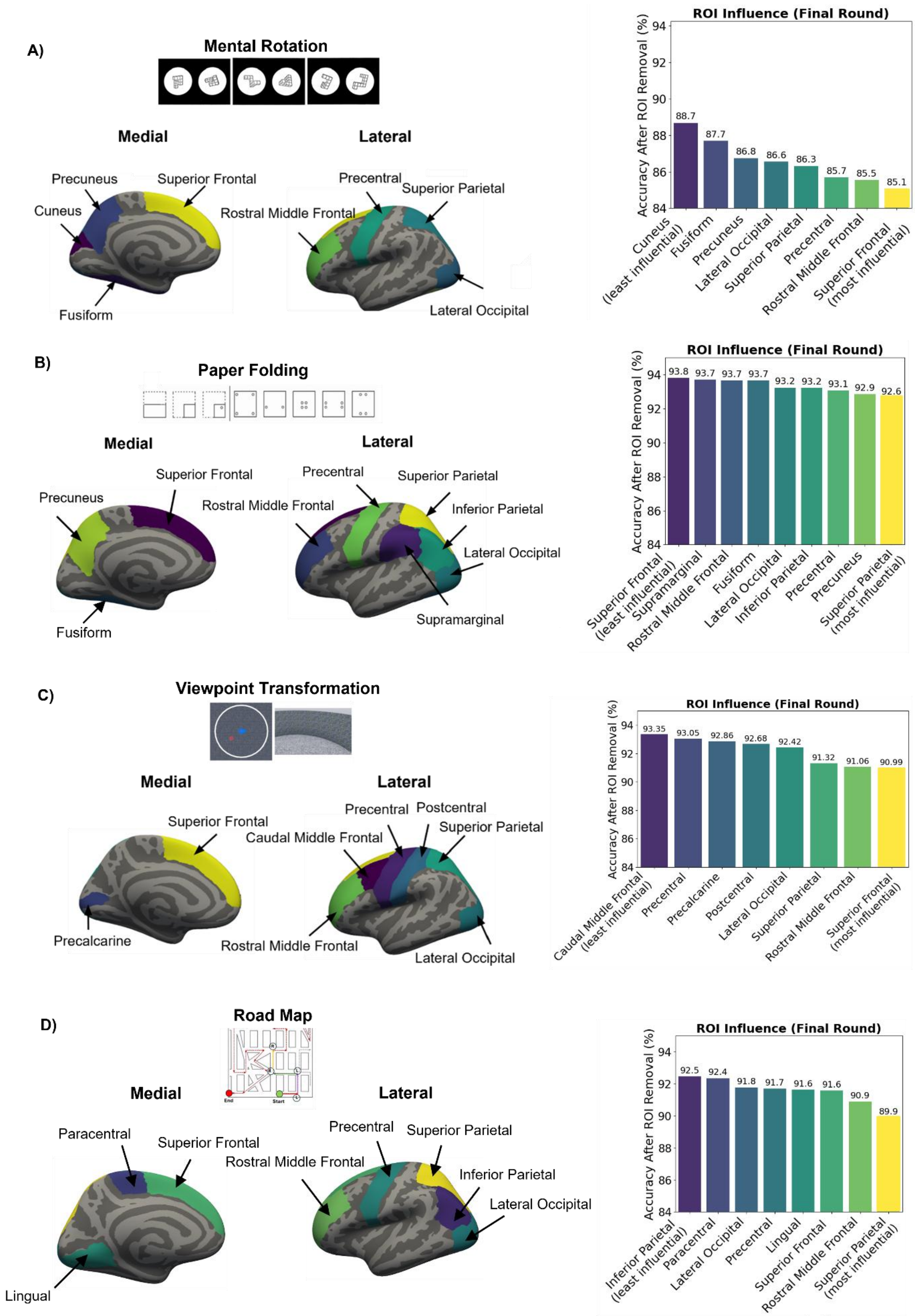
A)
Mental Rotation
Medial
Lateral
Precuneus
Superior Frontal
Cuneus
Fusiform
Precentral
Superior Parietal
Rostral Middle Frontal
Lateral Occipital
ROI Influence (Final Round)
Accuracy After ROI Removal (%)
88.7
87.7
86.8
86.6
86.3
85.7
85.5
85.1
Cuneus (least influential)
Fusiform
Precuneus
Lateral Occipital
Superior Parietal
Precentral
Rostral Middle Frontal
Superior Frontal (most influential)
B)
Paper Folding
Medial
Lateral
Superior Frontal
Precuneus
Fusiform
Precentral
Superior Parietal
Rostral Middle Frontal
Inferior Parietal
Lateral Occipital
Supramarginal
ROI Influence (Final Round)
Accuracy After ROI Removal (%)
93.8
93.7
93.7
93.7
93.2
93.2
93.1
92.9
92.6
Superior Frontal (least influential)
Supramarginal
Rostral Middle Frontal
Fusiform
Lateral Occipital
Inferior Parietal
Precentral
Precuneus
Superior Parietal (most influential)
C)
Viewpoint Transformation
Medial
Lateral
Superior Frontal
Precalcarine
Precentral
Postcentral
Superior Parietal
Caudal Middle Frontal
Rostral Middle Frontal
Lateral Occipital
ROI Influence (Final Round)
Accuracy After ROI Removal (%)
93.35
93.05
92.86
92.68
92.42
91.32
91.06
90.99
Caudal Middle Frontal (least influential)
Precentral
Precalcarine
Postcentral
Lateral Occipital
Superior Parietal
Rostral Middle Frontal
Superior Frontal (most influential)
D)
Road Map
End
Start
Medial
Lateral
Paracentral
Superior Frontal
Lingual
Precentral
Superior Parietal
Rostral Middle Frontal
Inferior Parietal
Lateral Occipital
ROI Influence (Final Round)
Accuracy After ROI Removal (%)
92.5
92.4
91.8
91.7
91.6
91.6
90.9
89.9
Inferior Parietal (least influential)
Paracentral
Lateral Occipital
Precentral
Lingual
Superior Frontal
Rostral Middle Frontal
Superior Parietal (most influential)

**Figure 6. Cortical regions most influential for predicting performance across four spatial tasks.** Backward-elimination analysis identified the brain areas whose removal produced the largest declines in pairwise prediction accuracy. **A)** Mental Rotation: influential ROIs included (in order from most to least influential) superior parietal, rostral middle frontal, precentral, suoperior parietal, lateral occipital, precuneus, fusiform, and cuneus. **B)** Paper Folding: predictive regions included superior parietal cortex, precuneus, precentral, inferior parietal, lateral occcipital, fusiform, rostral middle frontal, supramarginal, and superior frontal, ordered from most to least influential. **C)** Viewpoint Transformation: key regions involved superior frontal, rostral middle frontal, superior parietal, lateral occipital, postcentral, pericalcarine, precentral, and caudal middle frontal, ordered from most to least influential. **D)** Road Map: Predictive regions included superior parietal, rostral middle frontal, superior frontal, lingual, precentral, lateral occipital, paracentral, and inferior parietal, ordered from most to least influential based on the magnitude of accuracy decline following removal. For each task, right-panel bar plots show the accuracy after removing each ROI in the final elimination round; larger accuracy drops indicate greater regional importance.

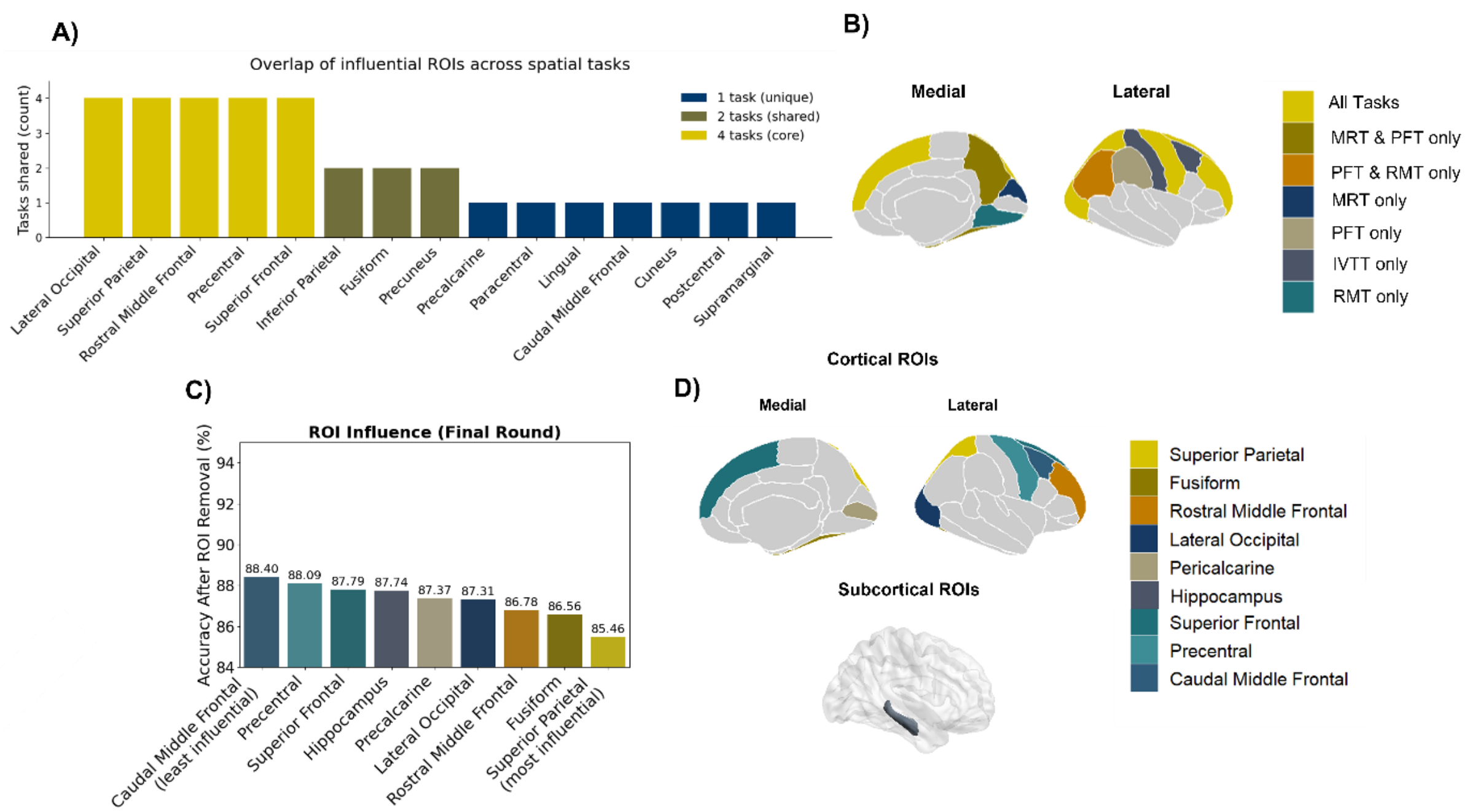

**Figure 7 | Overlap of cortical regions across spatial tasks**. **A)** Bar plot showing the number of spatial tasks in which each cortical region was identified as influential. Precentral, lateral occipital, superior frontal, rostral middle frontal, and superior parietal cortices were shared across all four tasks. **B)** Cortical surface maps illustrating overlap among regions identified for Mental Rotation (MRT), Paper Folding (PFT), Viewpoint Transformation (iVTT), and Road Map (RMT). Regions common to all four tasks are shown in light yellow. Precuneus was shared between MRT and PFT, but not iVTT or RMT. Inferior parietal lobe was uniquely shared between PFT and RMT. No regions were uniquely shared between iVTT and any other task. Task-specific regions are also shown. Together, these results indicate that object-based spatial transformations rely on a more tightly shared cortical substrate, whereas viewpoint transformation engages partially distinct structural systems. **C)** We tested the method using a very different cognitive paradigm, the Chronic Stress Scale. Bar plot shows accuracy after removal of each ROI in the final elimination round for predicting scores on the Chronic Stress Scale. Lower accuracy following removal indicates greater regional importance. The emergence of the hippocampus as an influential region

is consistent with its known role in chronic stress, supporting the validity of the method. **D)** Informative cortical and subcortical ROIs identified for chronic stress.

## Discussion

In this study, we asked whether a novel machine learning method could successfully predict performance on several spatial cognition tasks – using only anatomical T1 images as input – and provide insight into the brain regions that are important to predicting individual variability. Critically, we conducted this analysis on a relatively small dataset (108 participants), which for most machine learning algorithms would lead to catastrophic collapse. The results demonstrate that a pairwise ranking framework, combined with backward ROI elimination, can robustly decode inter-individual differences in spatial transformation abilities from structural MRI, revealing structure–behavior relationships that are difficult to recover with standard regression approaches in small, noisy datasets. This performance did not arise from model scale or overparameterization. Instead, it reflects the inductive capabilities inherent to convolutional architectures when trained on comparative rather than absolute behavioral targets. By reframing behavioral prediction as a relative decision problem, the model leverages a more stable supervisory signal that is less affected by noise, score discretization, and subject-specific measurement variability, revealing structure–behavior relationships that remain inaccessible to standard regression approaches.

Prior work has largely emphasized absolute score prediction, adopting linear models, kernel methods, or deep regression networks to link brain morphology to cognitive outcomes [24], [25], [26]. These approaches typically explain only modest variance and their predictive accuracy is constrained by the reliability of behavioral measurements, particularly in smaller samples [27]. Although deep learning can outperform classical models in large datasets, its advantages may be limited in neuroimaging studies when available sample sizes are insufficient to exploit nonlinear relationships [28]. Our findings show that pairwise supervision provides a principled alternative to direct regression. Rather than predicting an unstable scalar value, the network learns to solve the more biologically tractable problem of determining which of two individuals performs better. This reframing increases the effective dataset size quadratically, stabilizes optimization, and aligns model learning with well-established comparative mechanisms in human perception and decision-making.

Our interpretability results highlight the challenge of understanding CNN models in neuroimaging. Rather than relying on fine-grained attribution at the voxel level, we used backward ROI elimination to assess how removing anatomically defined regions affected model performance. This approach produced stable and reproducible profiles of important brain regions that aligned with known visuospatial networks, suggesting that spatial transformation ability depends on distributed cortical contributions spanning dorsal parietal, lateral occipital, and superior frontal regions.

Functional neuroimaging studies consistently implicate posterior parietal regions, including superior and inferior parietal cortex and the intraparietal sulcus, in visuospatial manipulation and mental rotation tasks [18], [19], [29]. Perspective-taking tasks, particularly those involving imagined changes in self-orientation, are often discussed in relation to parietal systems because they require the manipulation of spatial relationships and the updating of viewpoints [30]. Prior work also suggests that both object-based mental rotation and mentally taking the spatial perspective of another person (also called level-2 perspective taking) recruit lateral prefrontal regions in addition to superior and inferior parietal areas, consistent with roles for executive control and working memory in spatial transformation [28]. Gunia et al. [27] further suggest that perspective-taking tasks may involve a broader set of regions related to body- and self-representation, whereas object-based mental rotation more strongly engages visuospatial and lateral prefrontal systems. Together, these findings suggest that object-based and perspective-taking abilities rely on overlapping but partially distinct neural systems, and that the regions associated with individual differences may not necessarily be identical to those generally engaged during task performance.

In the present study, this pattern was also reflected in the regional profile identified by backward elimination. Superior parietal cortex emerged for RMT and iVTT, and inferior parietal cortex emerged for RMT, whereas the precuneus was observed primarily for the two object-based transformation tasks. The iVTT showed a somewhat different pattern, including postcentral involvement, which may indicate that different spatial tasks recruit partially distinct configurations of regions even when they are broadly classified under the same spatial domain. More broadly, our findings were not limited to parietal cortex: frontal regions also emerged across tasks, suggesting that individual differences in spatial performance depend on interactions among parietal, frontal, occipital, and sensorimotor systems rather than on a single anatomical locus.

These regional findings are also conceptually consistent with the cognitive demands of spatial transformation. Spatial ability tasks such as mental rotation and related transformation judgments are not thought to rely on a single focal region, but rather on coordinated processing across parietal, occipital, and frontal systems [31]. The prominence of parietal regions is likewise plausible, as superior and inferior parietal areas have long been implicated in spatial manipulation and coordinate transformations, while the precuneus has been linked to visuospatial imagery and the maintenance of spatial relations [19], [32], [33]. The recurrent involvement of precentral regions across tasks is also notable, as motor-related cortex has been implicated in spatial transformation and mental rotation, possibly reflecting embodied simulation (i.e., imagining using one's hands to rotate the object) or action-related coding during the manipulation of spatial representations [34], [35], [36], [37]. The involvement of lateral occipital cortex is likewise sensible given its role in visual shape and object representation [38], [39], which may support performance on tasks involving the analysis of complex spatial forms. Frontal contributions are also not surprising because successful performance requires maintaining spatial information and coordinating internal spatial representations. Taken together, the distributed profile recovered by backward elimination fits the view that spatial transformation ability emerges from interactions among perceptual, spatial, and executive systems rather than from a single facet of cognition.

Equally important, the credibility of the regional results depends on whether the model improves behavioral prediction relative to standard approaches. In our data, the pairwise framework outperformed conventional absolute-score models, including ridge, elastic net, and pointwise CNN regression models, which showed weaker generalization to unseen participants (lower accuracy, higher MAE). This comparison is critical because there is no voxel- or ROI-level ground truth for "importance" in structural MRI. The most direct validation available is therefore behavioral: if a model predicts individual differences in task performance more accurately and more robustly than standard baselines, then its derived regional importance patterns become more persuasive. In that sense, the pairwise framework is valuable not only because it yields interpretable ROI rankings, but because those rankings are produced by a model that better captures the behavior of interest than current standard methods.

Our findings also have implications for the broader debate on model complexity and data requirements in cognitive neuroscience. The strong performance achieved by a relatively compact convolutional architecture—when paired with pairwise supervision—demonstrates that high predictive accuracy does not require massive datasets or heavily engineered networks. Instead, appropriate inductive biases and problem formulations can dramatically boost data efficiency, allowing models to extract meaningful cognitive signatures from modest samples. This parallels observations in vision neuroscience, where comparative or contrastive learning objectives often reveal representational structure more effectively than absolute prediction [40].

Differences in generalization across behavioral tasks suggest that the effectiveness of pairwise prediction depends partly on the strength and reliability of the underlying brain–behavior relationship. MRT and iVTT showed comparatively strong generalization, whereas PFT and RMT exhibited larger discrepancies between training and held-out performance, indicating greater overfitting. One possibility is that some spatial measures capture more stable individual differences that are more strongly reflected in structural anatomy. Mental rotation, for example, is a well-characterized spatial ability with robust individual differences and may therefore provide a comparatively stable phenotype for brain–behavior prediction. Similarly, the iVTT yields a continuous position-error measure across repeated trials and requires coordinated perspective transformation and spatial updating, which may provide a sensitive measure of individual spatial ability. In contrast, PFT and RMT performance may be influenced more strongly by heterogeneous strategies, task-specific experience, measurement noise, or cognitive processes that are not well captured by structural MRI alone. The comparatively strong MRT performance may also partly reflect properties of the behavioral outcome itself. The MRT performance index combined accuracy and response time across 60 trials, capturing both the correctness and efficiency of mental rotation and potentially providing a richer and more stable representation of individual differences than measures based on a single behavioral dimension. These factors could affect the amount of reproducible signal available to the model and help explain why some tasks generalized better than others. Thus, although pairwise learning can mitigate the tendency of direct regression models to collapse toward the mean, its out-of-sample performance is likely to depend both on the strength of the underlying neuroanatomical signal and on the reliability and information content of the behavioral phenotype being predicted.The RMT analysis across two independent datasets also provides insight into the generalizability of the neuroanatomical features identified by the model. Although the exact set of informative regions differed between samples, superior parietal, inferior parietal, and paracentral regions were retained in both datasets. This partial overlap suggests that backward elimination may identify a combination of reproducible task-related signals and sample-dependent features. Differences in the remaining regions could reflect sampling variability, correlations among anatomical measures, differences in behavioral strategies across participants, or multiple anatomical features carrying partially redundant predictive information. Thus, the broader neuroanatomical systems associated with a spatial ability may generalize more consistently across samples than the exact set of ROIs selected in any single dataset. Additional independent datasets will be needed to determine which regional contributions are robust across populations and which are more sample-dependent.

An important goal for future work is to extend the ranking framework to multimodal imaging and longitudinal designs. Cognitive abilities such as spatial transformation draw on both structural and functional resources, and combining modalities may reveal how morphological constraints interact with dynamic neural processes to produce individual differences.

Although our approach provides advantages over existing methods, some limitations remain. While ROI elimination identifies regions whose presence is necessary for high predictive performance, it does not establish sufficiency or capture nonlinear interactions between regions. In addition, this method can indicate that a given region is important to determining individual performance, but the interpretation of that contribution is still limited. The model output cannot provide us information about how specific aspects of brain structure contribute to abilities. Moreover, the recurrence of several regions across multiple tasks suggests that some identified areas may reflect shared or domain-general components of performance rather than specificity to a single spatial process, although their relative importance may still vary by task. A finer-grained parcellation could help resolve some of these questions, although it would also increase the time and computer power needed. Future work could also adapt multivariate perturbation strategies to address these challenges.

Although the present study focuses on methodological innovation, our findings suggest promising future avenues for translational neuroscience and diagnostic imaging. A backward-elimination pairwise framework could function as a rapid, interpretable screening tool for neurodegenerative disorders such as Alzheimer's disease. Current diagnostic workflows often require multi-hour clinical evaluations, cognitive testing, and invasive or high-cost biomarker procedures such as CSF assays or PET imaging. In contrast, a pairwise based framework would require only a single T1-weighted MRI—a routine component of neurology visits—and inference could be performed in under one second on standard clinical hardware.

Beyond clinical neurology, the exponential data-efficiency of pairwise supervision has broader implications for small-sample domains. The same principle allowed lightweight CNNs trained from scratch to achieve competitive age estimation on the UTKFace dataset with orders of magnitude fewer images than typically required [15]. This finding demonstrates that exponential comparison graphs can unlock high-performing deep regression models in settings where absolute labels are noisy, scarce, or difficult to acquire, conditions common in medical imaging, behavioral neuroscience, and other biological sciences.

In sum, our results reveal that reframing cognitive prediction as a pairwise ranking problem, combined with region-level elimination, enables both robust decoding of individual spatial cognition performance and interpretable identification of its structural substrates. This work highlights how aligning model design with the comparative nature of human cognition can yield biologically meaningful insights, even in small datasets, and illustrates a path toward scalable and interpretable machine intelligence for understanding individual differences in the human brain, for both healthy and clinical populations. Together, these findings indicate that pairwise-comparison-based learning may provide a general pathway for achieving state-of-the-art prediction under extreme data limitations, offering a bridge between machine learning scalability and real-world biomedical constraints.

## Methods

## Participants

112 participants (age range: 18-35, *M* age: 21.5, *SD*: 3.73, 64 female, 46 male) from the University of California, Irvine (UCI) agreed to undergo MRI scanning as part of a larger study on individual differences. Participants all gave signed informed consent to be part of the study in accordance with the UCI IRB. These individuals passed the eligibility screening and were scanned at UCI's Facility for Imaging and Brain Research (FIBRE), which houses a 3 Tesla Siemens Prisma scanner equipped with a 32-channel head coil. Of the 112 participants, four were excluded due to low-quality scans and excessive motion artifacts, resulting in a final sample of 108 participants for analysis. Due to missing behavioral data, the number of participants used for model training varied by task. Specifically, 106 participants were included for the Mental Rotation Task (MRT), 107 for the iVTT, and 105 for the Road Map Test (RMT), with all 108 participants included in the Paper Folding Task (PFT). 102 participants completed the chronic stress scale.

## Experimental Tasks

This study was part of a broader project that included 40 cognitive tasks and questionnaires, with several spatial tasks. In the present report, we focused exclusively on the tasks described here. In this study, we used four spatial tasks, including three desktop tasks: mental rotation (MRT), paper folding (PFT), road map (RMT), and one VR task: immersive viewpoint transformation (iVTT). Participants completed the behavioral tasks over the course of 5 sessions, each lasting 1-2 hours. MRT and iVTT were conducted in session 4, and PFT and RMT were conducted in session 2. We also analyzed a questionnaire on chronic stress (Chronic Stress Scale, CSS), which was completed in an optional 6$^{th}$ session. If participants completed all behavioral tasks they could also complete an MRI scan of basic structure and function. All sessions were completed within approximately 2-8 weeks.

### *Perspective Taking Tasks*

**Immersive Viewpoint Transformation Task (iVTT)** [41]**.** The iVTT includes spatial updating of one's position while trying to reach a target location (Figure 1A). The ambulatory virtual environment for this task is a virtual desert with a circular walled arena. Each trial includes a map presentation and a navigation phase. First, the participant is shown a map for 2 seconds, indicating the location and orientation of the participant at the start of the trial, and the location of their goal. Then, during the navigation phase, the participant is returned to the first-person view at the center of the arena. They must then turn and walk to where they think the goal is located, and click the trigger of the handheld controller when they think they have reached the target. Critically, the heading orientation of the participant on the map does not align with the actual facing direction of the participant. In order to successfully reach the goal, the participant must both transform the perspective from an overhead viewpoint to a first-person viewpoint and update their heading with respect to the orientation in the map. There are 24 trials in total. The primary outcome measure is the position error, which was defined as the distance between the participant's final location and the target's true location.

**Road Map Test (RMT)** [42]. In this test of perspective taking, participants are shown a meandering travel path on a map of a town (Figure 1B). Without turning the map, they indicate whether each turn is to the left or right. There are 32 turns in the route, and the participants have 20 seconds to respond. The main outcome measure is the total number of correct turns, resulting in a maximum possible score of 32. A higher score indicates better perspective-taking ability.

***Object-based Transformation Tasks***

**Paper Folding Test (PFT)** [43]. This task requires the ability to mentally manipulate a piece of paper and imagine how the hole punch shows up spatially when the paper is unfolded (Figure 1C). A piece of paper is shown folded various times, and one hole is punched through all folds of the folded paper. The participant must decide from 5 options which unfolded paper matches what the paper would look like after the hole-punch. There were 2 parts with 10 trials each, and a 3-minute time limit for each part. The main outcome measure was the total number of correct responses, resulting in a maximum possible score of 20. A higher score indicates better spatial ability.

**Mental Rotation Test (MRT)** [44]. This task tests spatial visualization and accurate mental rotation of objects. Two sets of three-dimensional cubed objects are displayed side-by-side on the screen, which are either exactly the same or are mirror images of each other (Figure 1D). The cubed objects are also rotated, so they are not in the same picture plane. Participants decide whether the figures are the same or different/mirror images. There are 60 trials and this test was untimed. The main outcome measure is the total number of correct responses, resulting in a maximum possible score of 60. Another commonly-used measure is reaction time for correct trials, with faster responses indicating more ability. To combine these into a single performance index, we calculated Z-scores for both accuracy and RT (based on correct trials only), and then averaged the RT Z-score and the accuracy Z-score. This way, higher scores reflect both higher accuracy and faster responding.

***Questionnaires***

**Chronic Stress Scale (CSS).** The Chronic Stress Questionnaire consists of 51 items about everyday situations such as financial issues, marriage, work, etc. [45]. Participants need to indicate to what extent these conditions apply to their real-life using a 3-point scale, "Not True, Somewhat True, Very True," each rated as 0, 1, and 2, respectively. Some examples of the conditions are "You have a long-term debt or loan" or "Too much is expected of you by others." The measure is the average rating of all conditions, with a range from 0-2.

**Displays and Equipment**

MRT and PFT, and RMT were built in PsychoPy and presented on a desktop computer. Mouse clicks and/or keyboard responses are recorded for all tasks developed in PsychoPy. Experimental sessions utilizing iVTT took place at the Center for Ambulatory Virtual Environment Research in Navigation (CAVERN) on the University of California, Irvine campus. A Virtual Environment (VE) was built to fit in CAVERN, yielding a maximum of 7m x 7m size environments. Environments were displayed using an HTC VIVE Pro Eye VR head-mounted display (HMD) with Dual OLED 3.5” diagonal display (1440 1600 pixels per eye or 2880 1600 pixels combined), a 90 Hz refresh rate, and a 110° field of view capable of delivering high-resolution audio through removable headphones. In addition to the HMD, the VR interface included two HTC VIVE wireless handheld controllers for interacting with the experiment and four HTC Base Station 2.0 infrared tracking sensors for large-scale open space tracking. The Virtual Environment (VE) for the viewpoint transformation task was built and rendered in Unity; the task was programmed in Unity using custom scripts and scripts modified from the Landmarks Unity package [46]. In this task, participants physically walk in the environment while wearing a HMD. The VE provides complexity and ecological validity, as well as vestibular and proprioceptive information, while maintaining experimental control over the navigational factors of interest.

**Neuroimaging Data Collection and Preprocessing**

The scanning protocol included a whole-brain structural T1-weighted image using an MPRAGE sequence with the following parameters: repetition time (TR) = 2300 ms, echo time (TE) = 2.38 ms, inversion time (TI) = 902 ms, slices = 240, resolution = 0.8 x 0.8 x 0.8 $mm^3$, flip angle = 8°.

T1 images were pre-processed as part of an FreeSurfer [47], [48] preprocessing pipeline, which included skull stripping. T1 images were not normalized. To ensure data quality, all T1 scans were evaluated using MRIQC, which provides both automated quality metrics and standardized visual reports. All scans underwent manual visual inspection to identify motion artifacts, inhomogeneities, or preprocessing failures. A high-resolution T2-weighted image, a diffusion-weighted scan, an fMRI resting-state scan, and task-based fMRI scans were also acquired during this session, but were not analyzed in this study.

**Pairwise Convolutional Neural Network**

***Overview of Pairwise Data Construction and Supervision***

To reduce sensitivity to behavioral noise and avoid scale-dependent regression, we formulated spatial ability prediction as a pairwise ranking problem. For a dataset of $N$ subjects, all unique subject pairs were generated, yielding $N(N-1)/2$ training comparisons (4,950 pairs for 108 subjects). Each training example consisted of two preprocessed T1-weighted MRI volumes $(x_i, x_j)$ and a binary label (0,1) indicating which subject had the higher behavioral score. This transformation increased the effective number of training samples by two orders of magnitude relative to single-subject regression. The optimization was performed using cross-entropy loss on these pairwise targets.

Each ordered pair of subjects (i,j) was assigned a binary ranking label according to their behavioral scores,

$$y_{ij} = \begin{cases} 0, & x_i > x_j \\ 1, & x_i \leq x_j \end{cases}$$

where $(x_i)$ and $(x_j)$ denote the behavioral scores of subjects (i) and (j), respectively. The network therefore learned to predict which of the two subjects exhibited higher behavioral performance rather than estimating an absolute score directly.

We operated on 3D T1-weighted volumes, slicing each brain into 2D sections and applying augmentation. Two brain scans are independently encoded by shared slicing-network weights. Their resulting feature maps (one channel each) are stacked to form a two-channel tensor that is fed into the ranking network, which outputs a binary decision indicating which subject has higher behavioral performance (Figure 3A). Details of each step in the procedure are outlined below.

### ***Data Augmentation and 8× Test-Time Augmentation***

To ensure symmetry and increase robustness, each pair was additionally augmented by reversing input order $(x_j, x_i)$, doubling the dataset. A sagittal left–right flip was applied with 50% probability independently to each subject in the pair, yielding another 4× augmentation of each comparison. Across all augmentations, the resulting dataset contained approximately 160,000 training examples.

### ***Model Architecture***

We developed the pairwise regression framework on our dataset consisting of 108 T1-weighted MRI brain anatomical volumes and a continuous behavioral score for each subject. Each of the four spatial tasks were modeled separately, as was the Chronic Stress Scale. The model is a fully 2D convolutional architecture (minus the final fully-connected layer) applied slice-wise to sagittal MRI volumes. Each $192^3$ MRI scan was down-sampled to $96^3$ using trilinear interpolation for computational efficiency, while the corresponding FreeSurfer ROI labels were resampled using nearest-neighbor interpolation to preserve discrete anatomical labels. Background voxels were set to zero after preprocessing. During ROI-ablation experiments, voxels belonging to excluded ROIs were masked to zero while all remaining anatomy was preserved. The network consists of two stages: a slicing network that encoded each full 3D brain volume slice-by-slice, and a ranking network that performed the pairwise decision (Figure 3A).

Together, these components form a lightweight architecture roughly comparable to a ResNet-14 in depth, but structured around per-slice encoding rather than full 3D convolution. The network starts by passing each 3D brain scan into the shared slicing network one slice at a time. Once both brains have been sliced and turned into individual 2D feature maps, they are stacked on top of each other and fed as input to the smaller ranking net. The ranking net is trained to predict which subject has the higher behavioral score based on these paired representations.

Both the slicing and ranking networks were constructed from residual blocks containing two (3×3) convolutional layers followed by batch normalization and ReLU activation. Residual shortcuts used (1×1) convolutions whenever spatial resolution or channel dimensionality changed. The slicing network consisted of five residual blocks with adaptive average pooling to produce one feature vector per slice, whereas the ranking network contained two residual blocks followed by adaptive average pooling and a fully connected classifier producing the pairwise decision.

**Slicing Network and Feature Encoding.** The pairwise network was a 14-layer convolutional architecture inspired by ResNet blocks [49], applied to axial slices of size 96 × 96. The slicing network composed of 10 convolutional layers (a ResNet-10–style stack) that processes each sagittal slice independently to produce a per-slice feature embedding. For each subject, slice features were aggregated into a compact single-channel 2D feature representation (Figure 3B). To ensure a high ratio of data points to trainable parameters, the network was fully convolutional except for the final layer.

To improve convergence and reduce the impact of ROI masking in later elimination rounds, the slicing network was pretrained jointly with the ranking network using the full set of ROIs. After convergence, slicing features for all subjects were cached to enable rapid loading during subsequent backward elimination rounds.

**Pairwise Ranking Network.** The ranking network is composed of 4 convolutional layers plus a small fully connected layer, which receives the resulting slice-embedding map and outputs a single continuous score (Figure 3B). The ranking module received a stacked pair of encoded brain volumes and consisted of four convolutional layers with dropout (p = 0.3), followed by a fully connected layer producing a two-dimensional output representing the probability that subject $i$ or subject $j$ performed better. Directionality was preserved because the two input channels corresponded to the two subjects, and the convolutional filters were not constrained to be symmetric. Reversing subject order thus produced distinct feature representations, enabling the augmentation.

### *Model Training and Backward Elimination of Anatomical Features*

To obtain anatomically-grounded interpretability of the CNN, we systematically removed ROIs to assess how the removal affected model prediction accuracy. In each iteration, one unique ROI was removed, and the model was retrained on the remaining ROIs. ROIs whose removal did not reduce model accuracy were identified as non-informative. In each elimination round, the lowest one-third of ROIs ranked by impact on model accuracy were discarded, and the remaining ROIs were carried forward to the next round. The model was retrained across multiple iterations within each round until the ranking of ROI importance stabilized, defined as reaching the same ordering of ROIs in consecutive iterations based on their average impact on model accuracy.

For each round of ROI elimination, 80% of subjects were used for training and 20% were held out for evaluation. Early stopping based on validation accuracy prevented overfitting. Despite the small number of unique subjects, the large number of augmented pairwise comparisons enabled stable optimization; even with as few as 50 training and 50 test subjects, performance remained consistent, indicating that the model was not memorizing individual anatomy.

All models were trained using the cached slicing features to accelerate training. Approximately 10,000 ranking networks were trained across all ROI elimination rounds for the 96 × 96 × 96 resolution experiment. Training was performed on a workstation equipped with an NVIDIA RTX 2070 and GTX 980 Ti.

Models were optimized using the AdamW optimizer with a learning rate of ($7\times10^{-5}$) and weight decay of ($1\times10^{-2}$). Optimization minimized the binary cross-entropy loss between predicted and observed pairwise labels. Training was terminated using early stopping when held-out accuracy failed to improve for six consecutive epochs. During training, each subject in a pair underwent independent left-right flipping with probability 0.5 to improve robustness.

For a predicted probability $p_{ij}$ that subject $i$ outperformed subject $j$, optimization minimized the binary cross-entropy objective:

$$L = -y_{ij} \log p_{ij} - (1 - y_{ij}) log(1 - p_{ij})$$

where $y_{ij}$ denotes the ground-truth pairwise label, and $p_{ij}$ is the predicted probability that subject $i$ outperformed subject $j$.

**ROI Definition and DK Atlas.** Structural regions of interest were defined using the FreeSurfer *aparc+aseg* parcellation (aparc+aseg.nii). This atlas includes the Desikan–Killiany (DK) cortical parcellation, comprising 34 cortical regions per hemisphere (68 total cortical regions), as well as automated FreeSurfer segmentation of 42 subcortical structures. In total, the atlas contains 110 anatomical regions. For analysis, homologous left and right hemisphere regions were collapsed, and some parcellations were not bilateral regions (e.g. brainstem), resulting in 60 unique regions for analysis.

**Training on Random ROIs.** To evaluate the reliability of the backward elimination framework, we compared model performance when training on the informative ROIs identified by backward elimination versus randomly selected ROIs. This analysis tested whether the selected informative ROIs provided predictive value beyond what would be expected from an arbitrary set of regions. We trained 600 models using randomly selected sets of eight ROIs to generate a null distribution and six models using the eight informative ROIs. Models trained on randomly selected ROIs showed lower accuracy than models trained on the informative ROIs, suggesting that the backward elimination procedure reliably identifies regions that contribute meaningfully to predictive performance (See Supplementary Materials).

***Behavioral Regression and Decoding via Entropy***

At test time, a subject's score is inferred by aggregating pairwise predictions against all other subjects. The reference pool includes subjects from both the training and test sets, and reference subjects are grouped according to their observed behavioral scores**.** To visualize how the model reasons about the behavioral scores, we first examined a single test subject and its pairwise comparisons against reference subjects grouped into score bins. The model's pairwise predictions vary smoothly with score difference: pairs with similar scores produce probabilities near 0.5 and high entropy, whereas pairs with large score gaps saturate toward 0 or 1. Figure 4B illustrates the inferred score of the test subject, defined by the point at which the pairwise comparison probability (entropy) is 0.5, indicating maximal uncertainty about which subject scored higher.

Decision uncertainty was quantified using entropy:

$$H(p) = -p \log p - (1-p)log(1-p)$$

where *p* is the predicted probability that the test participant outperformed the reference participant. Maximum entropy occurs when ($p \approx 0.5$), indicating that the model is maximally uncertain because the two participants are estimated to have similar behavioral performance.

To improve robustness against local fluctuations, the probability curve was smoothed using a one-dimensional Gaussian filter ($\sigma = 1$) prior to entropy computation. This produced a stable estimate of the behavioral score while preserving the monotonic relationship between pairwise probability and behavioral similarity. The predicted behavioral score was therefore defined as the reference score corresponding to the maximum entropy value after Gaussian smoothing of the probability curve.

**Conventional Regression Baselines.** We also evaluated ridge regression and elastic net as feature-based baselines for predicting behavioral scores from regional cortical-thickness measures. To match the evaluation framework used throughout the study, 8 test subjects were selected across the behavioral score distribution, and the remaining subjects were used for training. Performance was assessed using mean absolute error (MAE) and $R^2$on both training and held-out test subjects (see Supplementary Materials).

***Pointwise CNN Regression Baseline.*** To determine whether the observed performance was specific to the pairwise learning formulation, we trained conventional pointwise CNN regression models for each behavioral task. The models used the same MRI preprocessing, subject split, and ResNet-style feature-extraction framework – including data augmentation – but directly predicted each participant's continuous behavioral score using a linear regression output and mean-squared error loss (see Supplementary Materials).

### *Application of Method to Independent Dataset*

Whole-brain T1-weighted structural MRI scans were acquired on a Siemens 3T Magnetom Prisma scanner at the UCSB Brain Imaging Center. T1-weighted MPRAGE images were collected at 0.9 mm isotropic resolution (TR = 2500 ms, TE = 2.22 ms, flip angle = 7°). A total of 102 participants from this independent study completed the same Road Map Test, although it was conducted on paper and pencil rather than on a computer. We applied the pairwise CNN framework to predict Road Map performance from T1-weighted MRI scans using entropy-based regression. We also used backward ROI elimination to identify regions that contributed most strongly to predictive performance.